\documentclass{article}

\usepackage[preprint]{neurips_2026}

\usepackage[utf8]{inputenc}
\usepackage[T1]{fontenc}
\usepackage{hyperref}
\usepackage{url}
\usepackage{booktabs}
\usepackage{amsmath}
\usepackage{amsfonts}
\usepackage{amssymb}
\usepackage{nicefrac}
\usepackage{microtype}
\usepackage{xcolor}
\usepackage{tabularx}
\usepackage{longtable}
\usepackage{graphicx}
\usepackage{wrapfig}
\usepackage{tikz}
\usepackage{enumitem}

\newcolumntype{Y}{>{\raggedright\arraybackslash}X}

\title{Evaluating Open-Weight E-Commerce Agents with Environment-Grounded Verification}

\author{%
  Nimit Shah \\
  AION
  \And
  Haitz Sáez de Ocáriz Borde \\
  AION
}

\begin{document}

\maketitle

\begin{abstract}
A shopping conversation has many routes to the same cart, and a task-success rate reduces all of them to one score. We build a deterministic and reproducible e-commerce environment that precommits each trial's customer and trajectory parameters, including the persona, difficulty, target cart, and an item reveal schedule. A simulated consumer attempts to buy a target cart from the environment with assistance from the evaluated model. The environment guides the simulator's actions and records every assistant action alongside the environment state at that point. After the trial, these records allow the evaluator to assess individual parts of the conversation against the retained evidence. For example, the evaluator penalizes a search for failing to surface a target product only when the customer has already mentioned that product. We further use this evidence to apply different penalties to tool calls depending on how the assistant's actions compare with an expected tool-call set. Our environment also interacts with the simulator bidirectionally, reading its output to stop the trial when the simulator determines that the customer has become too frustrated and injecting directives in real time that specify when to explore, defer buying an item, or recall a previous exchange. This interaction creates an open-ended and verifiable simulation. Across eight open-weight agents from 20B to 35B parameters, with 160 trials per agent and 44 metrics, the resulting capability profiles distinguish under-action, over-purchase, unsupported product attributes, and poor search, all of which terminal success obscures.
\end{abstract}

\section{Introduction}

Tool-using agents for online shopping must translate incomplete preferences into catalog actions while maintaining a cart the customer accepts. A final cart establishes the endpoint of that interaction, but it does not establish the route that produced it. The completed transcript alone cannot determine the catalog facts available to the agent, the cart state after each action, or whether a customer had requested an item at the time of a search. These missing facts prevent an evaluator from distinguishing an appropriate action from an accidental success or an unnecessary one.

We propose an e-commerce simulation environment that constructs the evidence required for verification before a conversation begins. The generator creates a hidden \emph{target cart} from a $10{,}000$-product control catalog and commits to a \emph{reveal schedule} that determines when each item becomes relevant to the customer. A controller applies that schedule during the conversation while a user simulator plays the customer. The environment retains the target cart, catalog state, tool trace, and reveal spans throughout the trial. This construction keeps the customer interaction open-ended while giving the evaluator a stable account of what the agent needed to know and do.

The verifier uses this account to reconstruct cart state and evaluate catalog-grounded claims, search behavior, and tool use. A model-based judge remains responsible only for questions that require language, including whether an attribute claim is supported by a catalog row and whether a response addresses the customer's request. The evaluation therefore separates failures that a final-cart score combines. It can identify whether an agent failed to find a requested item, added an item the customer did not want, described a product without support, or spent its turns without advancing the purchase.

We evaluate eight open-weight agents from 20B to 35B parameters on 160 trials per agent. The resulting profiles expose differences in retrieval, cart management, and grounded descriptions that have different implications for deployment and repair. We also examine shopping pace across the generator's difficulty tiers using termination reasons and tool-call density.

We make five contributions.

\begin{enumerate}[leftmargin=1.4em,itemsep=1pt,topsep=2pt]
\item An e-commerce environment whose scenario generator emits the ground truth required for verification. Each scenario contains a hidden target cart, per-item reveal spans, and a control catalog whose category mix is fitted to aggregate retail data (Section~\ref{sec:env}).
\item A battery of $44$ metrics for shopping outcomes, tool use, and customer-support quality. Deterministic state reconstruction settles questions that do not require a judge, and a standard registry supports new metrics (Section~\ref{sec:signals}).
\item Cohort-specific scoring. Each trial kind carries its own rubrics and applicability map, allowing an evaluation to target a deployment without averaging incompatible tasks (Section~\ref{sec:signals}).
\item A verification ablation that identifies which questions can be settled from a final cart, a transcript, environment state, or the full simulator record (Section~\ref{sec:ablation}).
\item An eight-model study that reads agents as capability profiles and uses termination reasons and tool-call density to diagnose difficulty-tier outcomes (Section~\ref{sec:results}).
\end{enumerate}

\section{Related work}

WebShop \citep{yao2022webshop} gives the agent a complete shopping request at the start and scores how well the purchased product satisfies it. Its authors validate this scorer and separately analyze search activity, products visited, and example trajectories, but the benchmark does not publish standardized correctness labels for individual actions.

ShoppingBench, WebMall, WebArena, and Mind2Web evaluate shopping or browser agents from task specifications supplied up front \citep{wang2026shoppingbench,peeters2026webmall,zhou2024webarena,deng2023mind2web}. Conversational shopping benchmarks expose preferences incrementally. ShopperBench conditions a dual-agent interaction on persona profiles \citep{ling2026shopperbench}, while ChatShop reveals hidden attributes and options of one target product in response to clarification questions \citep{chen2024chatshop}. Neither structure precommits per-item request boundaries for state-grounded checks.

Process-oriented evaluation also moves beyond terminal success. AgentBoard measures progress over manually labeled subgoals, and Kirgis et al. recommend systematic log analysis to expose failures hidden by outcome scores \citep{ma2024agentboard,kirgis2026loganalysis}. Conversational, stateful benchmarks provide the closest process comparisons. ToolSandbox matches fixed, human-authored milestone and minefield graphs to per-turn state snapshots, imposing a partial temporal order over selected events \citep{lu2025toolsandbox}. $\tau$-bench compares the final database in a simulated retail interaction with an annotated goal state \citep{yao2025taubench}. Its dialogue records when information was uttered, but its scenario does not expose a precommitted, structured reveal schedule, so its published reward cannot distinguish a search issued before a request from the same search issued after it. COMPASS progressively reveals travel constraints through a dynamically updated simulator prompt and records the turn of full specification, but it does not assign each constraint a reveal turn before dialogue \citep{qin2025compass}. Our scenarios add the precommitted, request-relevant boundaries needed to grade intermediate tool calls.

Large language models are widely used to evaluate generated text \citep{zheng2023judging,liu2023geval}, but studies document position and length biases and imperfect success classification \citep{wang2024fair,dubois2024lengthcontrolled,lu2025agentrewardbench}. Claim-decomposition methods provide a complementary approach to factuality evaluation. FActScore checks atomic claims against a knowledge source, while FacTool combines claim extraction with task-specific tools \citep{min2023factscore,chern2023factool}.

\section{The Evaluation Environment}
\label{sec:env}

We fix only the evidence required for verification before each trial and leave the surrounding customer interaction open-ended.

\subsection{The storefront}

We evaluate agents against a held-out control catalog of $10{,}000$ products across $19$ categories. We approximate a general e-commerce storefront by averaging category shares from six retailers. This distribution provides a grounded prior over categories rather than an exact reproduction of any retailer. Appendix~\ref{app:catalog} gives the data sources, normalization procedure, and aggregate and per-retailer shares.

\paragraph{Tool interface.} The agent acts through eight tools. The verifier groups them into four registry categories when penalizing unnecessary calls; Appendix~\ref{app:toolcat} gives the tool mapping and weights. The retrieval backend encodes queries and product text with \path{sentence-transformers/all-MiniLM-L6-v2} and ranks normalized embeddings by cosine similarity \citep{reimers2019sentencebert}. Each trial uses a fresh cart.

\subsection{Scenario: what the customer wants, and who they are}

A trial pairs the agent under test with a \emph{user simulator}, a frontier model that plays the customer. One integer seed determines the scenario. General shopping, grocery, and the refusal probes in Appendix~\ref{app:probes} share this machinery, but the harness generates and aggregates them separately. The trial kind determines which categories the basket draws from and how the customer works through it.

\paragraph{Target cart.} The customer is assigned a hidden set of in-stock products with required quantities, built category-first so the basket fills the way a real one does. A difficulty tier sets \emph{breadth}, the number of catalog categories visited. Within a category we draw a price-weighted representative, a Gaussian over log-price centered on the category median so that mid-range items dominate, and that price sets \emph{depth}. Writing $B$ for the breadth and $p_j$ for the price of the representative drawn in slot $j$, the basket holds $m=\sum_{j=1}^{B}D_j$ distinct products, with
\[
  \begin{aligned}
    P(D_j = d) &= r_j^{\,d-1}\,(1-r_j),
    &
    r_j &= \max\!\Big(0.07,\; 1 - \tfrac{2}{1+c_j}\Big), \\
    c_j &= \min\!\big(6,\ \max(1,\ \mathrm{round}(10 - 3.3\log_{10} p_j))\big).
  \end{aligned}
\]
A five-dollar staple gives $r_j\approx0.71$ and a mean depth near $3.5$; a seven-hundred-dollar item gives $r_j=0.07$ and almost always resolves to a single unit. The tier controls breadth alone, so the two simple tiers draw one category and reach several products through depth, while the multi-item and complex tiers multiply the basket by visiting more categories. Quantities follow an offset power law $w(q)\propto (q+c)^{-\alpha}$ with two regimes. Setting $c=0$ leaves the head steep, so the shopper takes one unit, while a large $c$ flattens it, so one through five units are comparably likely. Groceries use the flat regime with a taper that keeps party-size counts rare. Appendix~\ref{app:quantity} gives the remaining parameters.

Because every goal item is drawn from the catalog, the customer never asks for something the store does not carry. An agent that reports no match has failed to retrieve rather than found an empty shelf, and Section~\ref{sec:signals} scores it that way.

\paragraph{Difficulty and persona.} Each trial carries a difficulty from $1$ to $10$ mapped to one of four tiers (Table~\ref{tab:tiers}). The turn budget is basket-aware rather than a fixed function of difficulty. It is $\text{slack} + \mathrm{round}(w_{\text{tier}} \cdot m)$ for $m$ realized distinct products, with slack $3$ and a per-tier weight growing from $1.0$ to $2.25$. Within a tier, raw difficulty $d$ determines whether the customer withholds optional hints. Each hint about size, color, fit, or budget is withheld with probability $d/10$. The customer is also drawn along six independent categorical dimensions with realistic-frequency weights. These dimensions specify how precisely the customer states what they want, how readily they commit, how much price matters, their product vocabulary, their message length, and their typing quality. Appendix~\ref{app:persona} lists the values.

\begin{table}[t]
  \caption{The four difficulty tiers. Breadth is the number of catalog categories the target cart spans. Grocery trials use the wider breadth ranges. The turn-budget weight $w_{\text{tier}}$ multiplies the realized basket size.}
  \label{tab:tiers}
  \centering
  \small
  \setlength{\tabcolsep}{3pt}
  \begin{tabularx}{\textwidth}{@{}l c >{\centering\arraybackslash}X c Y@{}}
    \toprule
    Tier & Difficulty & Breadth (general / grocery) & $w_{\text{tier}}$ & Customer behavior \\
    \midrule
    Direct purchase   & 1--2  & 1 / 1       & 1.00 & One item, commits to the first reasonable match. \\
    Light exploration & 3--5  & 1 / 1       & 1.25 & A few comparison questions, then settles. \\
    Multi-item         & 6--8  & 2--4 / 3--6 & 1.50 & A list; every item must be in the cart at the end. \\
    Complex            & 9--10 & 1--7 / 4--8 & 2.25 & Explores, defers, and circles back. \\
    \bottomrule
  \end{tabularx}
\end{table}

\subsection{The reveal schedule}
\label{sec:reveal}

Large target carts require the customer to express requests across several messages, making request timing part of the answer key. A search issued at turn $2$ cannot be held against an item the customer first mentions at turn $9$. We therefore fix the schedule in the scenario at build time and run it through a driver-side controller, which stamps each item with the customer-message index at which it is first requested.

Each multi-item trial draws one of three styles based on the realized basket and trial kind. \texttt{batch} states the list up front, \texttt{linear} releases one item at a time in order after the preceding item reaches the cart, and \texttt{nonlinear} follows a beat itinerary that can introduce and explore an item, defer it while pivoting to another request, and return later to settle it. These phases can interleave across items rather than follow basket order. Grocery is list-shaped rather than a journey, so it is batch-dominant and never nonlinear; non-grocery complex trials lean nonlinear. Appendix~\ref{app:reveal} gives the sampling weights.

A \texttt{batch} trial with more than four products instead releases a bounded wave of items and advances once every item in the current wave is in the cart. Ordinary waves group related items, while complex trials may mix categories. Appendix~\ref{app:reveal} gives the wave construction.

Every nonlinear goal item receives one \emph{essential} add beat, while optional beats control its introduction, exploration, deferral, and return. This construction keeps every target achievable while allowing the conversation to move among unresolved requests. Appendix~\ref{app:reveal-examples} shows recorded linear, micro-batch, and nonlinear conversations produced by these schedules.

\subsection{Running a trial}
\label{sec:loop}

A turn contains one customer message and one agent turn, which may span several tool-call rounds.

The simulator answers each turn with a single structured call. Its five fields contain a short rationale, an inline frustration rating, the next customer message, a termination flag with its reason, and the chain of tool calls it expects the agent to make in response to the message it is about to send. That last field is the reference set for tool-call correctness in Section~\ref{sec:signals}. The simulator sees the persona, the target cart, the transcript so far, and the agent's own system prompt verbatim. Providing the system prompt lets it tell a clarifying question the agent's policy mandates from one that wastes the customer's time.

The customer it plays, however, cannot see the tool messages. The simulator reads them to rate the turn and to decide whether the goal is met, and it voices the customer's needs in character rather than quoting or paraphrasing a search payload, a product identifier, a price, or a stock count. This rule keeps search-result knowledge out of customer messages, so the agent must locate catalog items from the requests it can see.

The controller reads the ground-truth cart after every agent turn rather than scraping the transcript for it. On a nonlinear trial it also owns termination, ignoring a premature ``done'' from the simulator until the cart genuinely holds every goal item at its required quantity. The six recorded termination reasons are goal completion, turn-budget exhaustion, customer abandonment, an agent failure, a simulator failure, and a missing opening message. Section~\ref{sec:results} reads that mix directly.

The verification signals that follow use the reveal spans defined in Section~\ref{sec:reveal}. Search metrics count a target only after the customer has requested it.

\begin{figure}[t]
  \centering
  \begin{tikzpicture}[
    x=0.01065cm,
    y=-0.01065cm,
    every node/.style={inner sep=0pt, outer sep=0pt, text=black!86},
    figure title/.style={font=\fontsize{9.8}{10.5}\selectfont\bfseries},
    section title/.style={font=\fontsize{9.0}{9.7}\selectfont\bfseries},
    label/.style={font=\fontsize{8.7}{9.4}\selectfont},
    card title/.style={font=\fontsize{8.3}{9.0}\selectfont\bfseries},
    multi title/.style={font=\fontsize{7.7}{8.4}\selectfont\bfseries},
    support/.style={font=\fontsize{7.7}{8.4}\selectfont, text=black!68},
    box/.style={draw=black!35, fill=white, rounded corners=2pt,
      line width=0.45pt},
    blue panel/.style={draw=blue!65!black, fill=blue!4,
      rounded corners=3pt, line width=0.7pt},
    orange panel/.style={draw=orange!78!black, fill=orange!5,
      rounded corners=3pt, line width=0.7pt},
    green panel/.style={draw=green!48!black, fill=green!4,
      rounded corners=3pt, line width=0.7pt},
    purple panel/.style={draw=violet!70!black, fill=violet!4,
      rounded corners=3pt, line width=0.7pt},
    dashed box/.style={draw=black!45, fill=white, rounded corners=2pt,
      dashed, line width=0.55pt},
    blue flow/.style={->, >=latex, draw=blue!65!black, line width=0.8pt},
    orange flow/.style={->, >=latex, draw=orange!82!black, line width=0.8pt},
    green flow/.style={->, >=latex, draw=green!48!black, line width=0.8pt},
    purple flow/.style={->, >=latex, draw=violet!72!black, line width=0.8pt},
    gray flow/.style={->, >=latex, draw=black!52, line width=0.65pt},
    dashed flow/.style={dashed}
  ]
  \node[figure title, anchor=base west] at (24,32)
    {Environment-grounded verification pipeline};

  \draw[blue panel] (20,48) rectangle (1260,238);
  \node[section title, anchor=base west] at (40,78)
    {Scenario construction: ground truth fixed before interaction};

  \draw[box] (37,104) rectangle (177,150);
  \node[label, anchor=base] at (107,134) {Seed};
  \draw[box] (37,160) rectangle (177,220);
  \node[multi title, anchor=base] at (107,184) {Kind +};
  \node[multi title, anchor=base] at (107,211) {difficulty};

  \draw[box] (202,104) rectangle (352,220);
  \node[multi title, anchor=base] at (277,153) {Control};
  \node[multi title, anchor=base] at (277,179) {catalog};

  \draw[blue panel] (382,112) rectangle (562,212);
  \node[multi title, anchor=base] at (472,154) {Scenario};
  \node[multi title, anchor=base] at (472,181) {generator};

  \draw[blue panel] (590,94) rectangle (1120,228);
  \node[section title, anchor=base] at (855,126) {Precommitted scenario};
  \draw[box] (610,140) rectangle (760,216);
  \node[multi title, anchor=base] at (685,160) {Target};
  \node[multi title, anchor=base] at (685,184) {cart};
  \node[support, anchor=base] at (685,209) {quantities};
  \draw[box] (780,140) rectangle (930,216);
  \node[multi title, anchor=base] at (855,160) {Reveal};
  \node[multi title, anchor=base] at (855,184) {schedule};
  \node[support, anchor=base] at (855,209) {+ timing};
  \draw[box] (950,140) rectangle (1100,216);
  \node[label, anchor=base] at (1025,160) {Persona};
  \node[support, anchor=base] at (1025,184) {+ turn};
  \node[support, anchor=base] at (1025,209) {budget};

  \draw[blue flow] (177,127) -- (378,127);
  \draw[blue flow] (177,193) -- (378,193);
  \draw[blue flow] (352,162) -- (378,162);
  \draw[blue flow] (562,162) -- (586,162);

  \draw[orange panel] (20,258) rectangle (1260,710);
  \node[section title, anchor=base west] at (40,294)
    {Trial execution and environment-grounded verification};

  \draw[orange panel] (30,320) rectangle (315,580);
  \node[card title, anchor=base] at (172,356) {Interaction loop};
  \draw[box] (43,382) rectangle (193,470);
  \node[label, anchor=base] at (118,405) {Reveal};
  \node[support, anchor=base] at (118,433) {controller};
  \node[support, anchor=base] at (118,459) {simulator};
  \draw[box] (205,390) rectangle (305,462);
  \node[label, anchor=base] at (255,435) {Agent};
  \draw[box] (106,510) rectangle (246,572);
  \node[multi title, anchor=base] at (176,534) {Tools +};
  \node[multi title, anchor=base] at (176,560) {store};
  \draw[orange flow] (193,408) -- (201,408);
  \draw[gray flow] (205,450) -- (197,450);
  \draw[orange flow] (246,463) -- (220,501);
  \draw[gray flow] (145,501) -- (128,471);

  \draw[blue flow] (1120,161) -- (1150,161) -- (1150,300) --
    (172,300) -- (172,310);
  \draw[blue flow, dashed flow] (277,220) -- (277,246) -- (25,246) --
    (25,541) -- (101,541);

  \draw[green panel] (320,312) rectangle (590,652);
  \node[card title, anchor=base] at (455,342) {Evidence};
  \draw[box] (328,370) rectangle (582,424);
  \node[card title, anchor=base] at (455,405) {Final cart};
  \draw[box] (328,438) rectangle (582,492);
  \node[card title, anchor=base] at (455,473) {Transcript};
  \draw[box] (328,506) rectangle (582,568);
  \node[multi title, anchor=base] at (455,532) {Environment};
  \node[multi title, anchor=base] at (455,558) {state};
  \draw[box] (328,582) rectangle (582,644);
  \node[multi title, anchor=base] at (455,608) {Simulator};
  \node[multi title, anchor=base] at (455,634) {record};
  \draw[green flow] (315,446) -- (320,446);

  \draw[purple panel] (595,312) rectangle (1025,652);
  \node[section title, anchor=base] at (810,344) {Verifier};
  \node[support, anchor=base] at (810,372) {strongest available evidence};
  \draw[box] (615,392) rectangle (1005,468);
  \node[label, anchor=base] at (810,420) {Evaluation questions};
  \node[support, anchor=base] at (810,448) {outcomes, actions, claims};
  \node[support, anchor=base] at (810,464) {tool use, experience};

  \draw[purple panel] (614,492) rectangle (740,584);
  \node[section title, anchor=base] at (677,522) {Rule};
  \node[support, anchor=base] at (677,550) {fixed};
  \node[support, anchor=base] at (677,576) {checks};
  \draw[purple panel] (747,492) rectangle (873,584);
  \node[section title, anchor=base] at (810,522) {Model};
  \node[support, anchor=base] at (810,550) {LLM};
  \node[support, anchor=base] at (810,576) {checks};
  \draw[purple panel] (880,492) rectangle (1006,584);
  \node[section title, anchor=base] at (943,522) {Hybrid};
  \node[support, anchor=base] at (943,550) {rule};
  \node[support, anchor=base] at (943,576) {+ model};
  \draw[dashed box] (678,600) rectangle (942,644);
  \node[label, anchor=base] at (810,619) {Infer};
  \node[support, anchor=base] at (810,641) {transcript only};

  \draw[purple flow] (590,456) -- (605,456);
  \draw[purple flow] (810,468) -- (810,482) -- (677,482) -- (677,488);
  \draw[purple flow] (810,468) -- (810,488);
  \draw[purple flow] (810,468) -- (810,482) -- (943,482) -- (943,488);
  \draw[gray flow, dashed flow] (1005,430) -- (1015,430) --
    (1015,622) -- (952,622);

  \draw[purple panel] (1040,334) rectangle (1245,634);
  \node[multi title, anchor=base] at (1142,364) {Capability};
  \node[multi title, anchor=base] at (1142,394) {profile};
  \node[support, anchor=base west] at (1058,438) {Outcomes};
  \fill[blue!65!black] (1195,421) rectangle (1231,437);
  \node[support, anchor=base west] at (1058,474) {Retrieval};
  \fill[green!48!black] (1195,457) rectangle (1224,473);
  \node[support, anchor=base west] at (1058,510) {Cart};
  \fill[orange!82!black] (1195,493) rectangle (1235,509);
  \node[support, anchor=base west] at (1058,546) {Grounding};
  \fill[violet!72!black] (1195,529) rectangle (1228,545);
  \node[support, anchor=base west] at (1058,582) {Tool use};
  \fill[blue!65!black] (1195,565) rectangle (1232,581);
  \node[support, anchor=base west] at (1058,618) {Experience};
  \fill[green!48!black] (1195,601) rectangle (1225,617);
  \draw[purple flow] (1025,470) -- (1030,470);
\end{tikzpicture}
  \caption{Environment-grounded verification pipeline. The precommitted scenario drives the interaction, whose retained evidence supports fixed (Rule), model-based (Model), and hybrid checks. Transcript-only inferences lack verification evidence; verified measurements form a capability profile.}
  \label{fig:verification-process}
\end{figure}
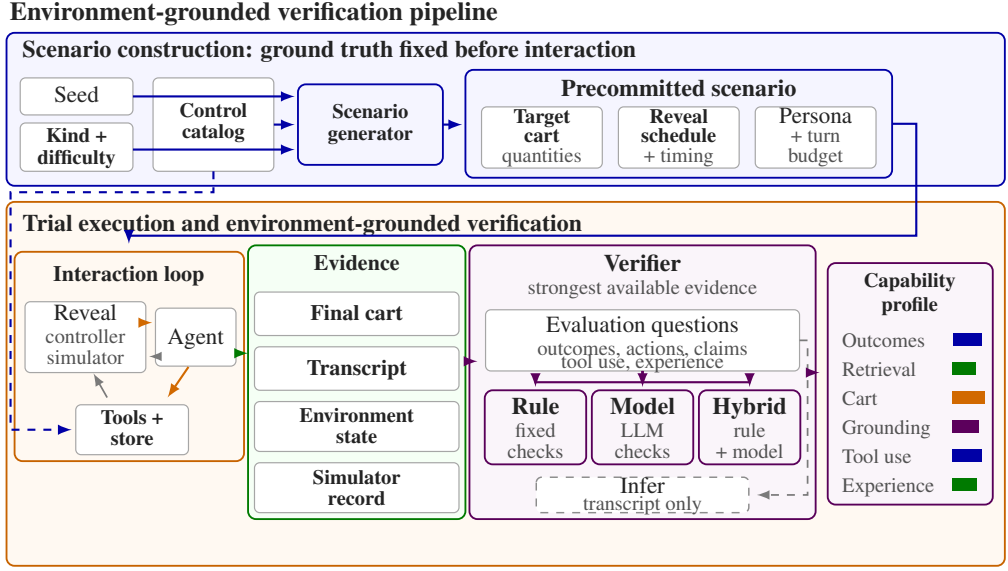

Figure~\ref{fig:verification-process} traces the trial from the precommitted scenario through the interaction loop and retained evidence to the verifier.

\section{Verification signals}
\label{sec:signals}

Verification uses the strongest evidence the harness has for each question. By \emph{state replay}, we mean reapplying successful tool results in transcript order to reconstruct the cart after each mutation. Together with the catalog and tool log, state replay settles what the agent retrieved and changed in the store state; a judge is used only for customer-facing meaning or quality that the record cannot settle.

\subsection{Deterministic signals}

The state replay reconstructs a final mapping from product to quantity, which the outcome metrics compare against the target. \textbf{Exact-cart success} requires the cart to hold exactly the target items at exactly the target quantities. We also report cart precision and recall to separate adding spurious items from missing wanted ones. With $m=\sum_{(p,q)\in G}\min(C[p],q)$ over target $G$ and reconstructed cart $C$, precision is $m/\sum_p C[p]$ and recall is $m/\sum_{(p,q)\in G} q$. \textbf{Goal achievement} is the weaker condition $C[p]\ge q$ for every target. \textbf{Goal progress} samples the fraction of exactly-satisfied target products after every cart mutation and takes the trial's maximum, so over-adding drops an item back out of the satisfied set.

For tool use we tag each call as \emph{errored} when its result reports an error or \emph{redundant} when its signature repeats with no intervening state change. We further tag searches as \emph{goal-relevant} when at least one target product is both missing and already revealed according to Section~\ref{sec:reveal}, \emph{empty} when a clean execution returns no results, and \emph{bad} when a goal-relevant search errors, returns nothing, or returns none of the still-needed targets. A bad search is \emph{recovered} when a later same-category retry succeeds, and \emph{autonomous} when that retry happens in the same turn without a customer re-prompt. The rates are $n_{\text{err}}/n_{\text{call}}$, $n_{\text{redun}}/n_{\text{call}}$, $n_{\text{empty}}/n_{\text{search}}$, $n_{\text{bad}}/n_{\text{rel}}$, and $n_{\text{rec}}/n_{\text{bad}}$.

\subsection{The need for model-based judging}

Two model-graded sources are kept apart. The user simulator rates one axis inline as the conversation unfolds, \textbf{user frustration}, reduced per trial to the worst turn. It reports the experience of the customer it plays. As frustration rises, the shopper grows terser and abandons sooner. Separately, an \textbf{LLM judge} per metric re-reads the finished transcript against written rubrics, writing an evidence-based rationale before emitting a score that we normalize to $[0,1]$ \citep{liu2023geval}.

The \textbf{hallucination} rubric applies this division within one response. Following claim-decomposition approaches \citep{min2023factscore,chern2023factool}, one structured call extracts each concrete claim from the agent's prose and types it as action, numeric, or attribute. The verifier labels an action claim a phantom action when the tool log contains no matching call. It checks numeric claims against catalog prices, stock, and the state-replayed cart. Only attribute claims reach the judge, which labels them as supported, embellished, contradicted, or ungrounded against the ground-truth catalog row. An ungrounded claim also detects a fabricated product. With $V$ the verifiable claims and $B\subseteq V$ the violations, the headline score is
\[
  H = 1 - \frac{|B|}{|V|}.
\]
The verifier reports the three channel rates separately. Table~\ref{tab:diagnostics} shows two agents differing fourfold on the attribute channel while agreeing on the numeric one.

The remaining rubrics assess whether each reply moved the purchase forward, whether the agent remembered and understood the customer's request, whether it asked for information it already had, whether recommendations were relevant, and whether it confirmed variant and quantity before adding to the cart. The clarification and helpfulness rubrics receive the agent's system prompt and treat policy-mandated clarification as compliant behavior.

\subsection{Tool-call correctness}

The verifier must distinguish a missed action from a reasonable additional lookup, while treating an unjustified cart mutation more seriously than an extra read. The reference set comes from the simulator, which predicts the immediate chain of calls the agent should make in response to each customer message. Let $E$ be that prediction and $A$ the calls actually issued. We group both by tool name and first apply the cheapest checks. A pair matches with no model call when its arguments are an exact JSON match or the tool takes no arguments. Only same-name pairs with differing arguments are queued for the judge, along with unmatched actual calls, so the judge decides whether a call was justified rather than spurious. All queued questions across the dialog are answered in one batched call, and assignment is greedy and order-stable so a rerun reproduces it. The tool category determines the penalty for surviving extras. Cart writes get no free allowance, read-only lookups are forgiven in bulk, and within a category the cost grows with each further extra. Writing $w_{\text{extra}}$ for the summed penalty, the per-turn score is $s_{\text{turn}} = \text{matched}\,/\,(|E| + w_{\text{extra}})$, so a miss shrinks the numerator and an unjustified extra inflates the denominator. Turns with neither expected nor actual calls are skipped. Appendix~\ref{app:toolcat} gives the category weights.

\subsection{Aggregation}

The harness reports shopping and grocery trajectory and conversational metrics separately. Each rubric declares its applicable trial kinds, and the runner skips inapplicable metrics rather than recording zero; Appendix~\ref{app:probes} explains how this applicability map handles refusal probes. Aggregates report the mean, sample standard deviation, and valid count, with Boolean means interpreted as rates.

\section{Verifiable simulation for e-commerce agents}
\label{sec:results}

Eight open-weight agents run the same $160$-trial sweep, $100$ general shopping, $50$ grocery, and $10$ refusal probes, across difficulties $1$--$10$, with Claude Sonnet 4.5 as both user simulator and primary judge \citep{anthropic2025sonnet45}. Every run uses the same tools, scoring contract, controller, and stopping policy, with model-specific native chat templates and parsers. The comparison is conditional on this configuration because context construction, tool mediation, and stopping rules are part of the experimental condition \citep{zhang2026harness}. For each agent, we resample its $100$ shopping trials with replacement $6000$ times and report the central $95\%$ of the resulting exact-cart success rates. For reported between-agent contrasts, we independently resample the two unpaired runs and take the central $95\%$ of the rate difference. We use the agents to test what the e-commerce environment can establish from its stored cart, catalog, reveal schedule, and tool trace. Appendix~\ref{app:probes} describes the refusal cohort, Appendix~\ref{app:latency} gives checkpoint identities, and Appendix~\ref{app:full} gives extended shopping metrics.

\subsection{Verification ablation}
\label{sec:ablation}

We test four retained-evidence conditions on each completed trajectory. Final cart contains terminal product quantities and the target comparison. Transcript contains the customer and assistant messages, executed tool calls, and results, but excludes the hidden target and other environment state. State adds the target cart, control catalog, and termination record, enabling state replay. Full simulator adds the reveal schedule, per-turn expected tool calls, and the user simulator's frustration record. For ten evaluation questions, Table~\ref{tab:information-ablation} labels each answer as Rule, Model, Hybrid, or Infer. Rule uses fixed verification, Model uses grounded output from the user simulator or judge, and Hybrid combines a model-derived signal with a fixed comparison or reduction. Infer lacks the retained evidence required for verification.

\begin{table}[t]
  \caption{Verification ablation across the four retained-evidence conditions defined in the text.}
  \label{tab:information-ablation}
  \centering
  \small
  \setlength{\tabcolsep}{3.5pt}
  \begin{tabularx}{\textwidth}{Y c c c c}
    \toprule
    Verification question & Final cart & Transcript & State & Full simulator \\
    \midrule
    Exact target cart & Rule & Infer & Rule & Rule \\
    Missing item versus over-purchase & Rule & Infer & Rule & Rule \\
    Progress before termination & -- & Infer & Rule & Rule \\
    Executed versus phantom action & -- & Hybrid & Hybrid & Hybrid \\
    Price, stock, and cart-total claim & -- & Infer & Hybrid & Hybrid \\
    Product-attribute support & -- & Infer & Model & Model \\
    Budget exhaustion versus abandonment & -- & Infer & Rule & Rule \\
    Customer frustration & -- & Infer & -- & Model \\
    Tool-call appropriateness & -- & Infer & Infer & Hybrid \\
    Search success for currently requested items & -- & Infer & Infer & Rule \\
    \bottomrule
  \end{tabularx}
\end{table}

Across fixed trajectories, the final cart establishes terminal outcome questions. With environment state, the verifier also establishes progress, executed actions, catalog-grounded claims, and termination reasons. With the full simulator record, it further checks request-relative tool appropriateness and search success and reports the user simulator's recorded frustration. Appendix~\ref{app:cases} gives three trial-level examples of these distinctions.

\subsection{E-commerce state turns outcomes into verifiable diagnoses}
\label{sec:profiles}

Exact-cart success ranges from $0.30$ to $0.57$, and the independently resampled contrasts separate the highest rate from the two lowest. The tier breakdown shows why these outcomes do not form a stable rank order. Gemma4-26B moves from $0.59$ on direct purchases to $0.27$ on light exploration and $0.53$ on complex baskets, while GPT-OSS moves from $0.59$ on direct purchases to $0.42$ on light exploration and $0.67$ on multi-item baskets. We therefore compare capability profiles rather than rank agents by one outcome. Appendix~\ref{app:pace} reports the full estimates and confidence intervals.

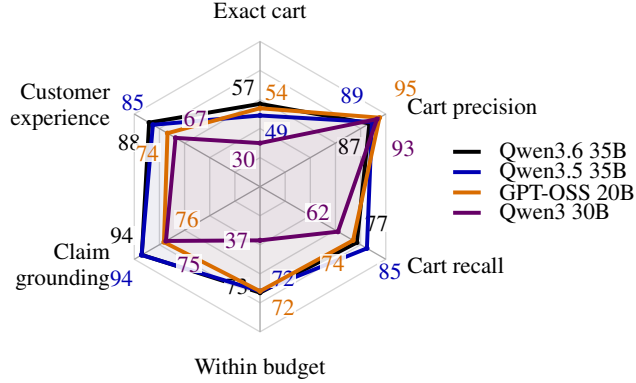
\begin{figure}[t]
  \centering
  \resizebox{0.62\textwidth}{!}{%
  \begin{tikzpicture}[
      scale=1.55,
      every node/.style={font=\scriptsize},
      profile value/.style={font=\scriptsize, fill=white, fill opacity=0.88, text opacity=1, inner sep=0.35pt}
    ]
    \foreach \r in {0.2,0.4,0.6,0.8,1.0} {
      \draw[black!16] (90:\r) -- (30:\r) -- (-30:\r) -- (-90:\r) -- (-150:\r) -- (150:\r) -- cycle;
    }
    \foreach \a in {90,30,-30,-90,-150,150} {
      \draw[black!35] (0,0) -- (\a:1);
    }
    \node[anchor=south] at (90:1.08) {Exact cart};
    \node[anchor=west] at (30:1.08) {Cart precision};
    \node[anchor=west] at (-30:1.08) {Cart recall};
    \node[anchor=north] at (-90:1.10) {Within budget};
    \node[anchor=east, align=right] at (-150:1.10) {Claim\\grounding};
    \node[anchor=east, align=right] at (150:1.10) {Customer\\experience};

    \draw[very thick, black, fill=black, fill opacity=0.035]
      (90:0.57) -- (30:0.87) -- (-30:0.77) -- (-90:0.73) -- (-150:0.94) -- (150:0.882) -- cycle;
    \foreach \a/\r in {90/0.57,30/0.87,-30/0.77,-90/0.73,-150/0.94,150/0.882} {
      \fill[black] (\a:\r) circle (0.018);
    }
    \node[profile value, text=black, anchor=south east, xshift=-1pt, yshift=2pt] at (90:0.57) {57};
    \node[profile value, text=black, anchor=north east, xshift=-2pt, yshift=-4pt] at (30:0.87) {87};
    \node[profile value, text=black, anchor=south west, xshift=2pt, yshift=3pt] at (-30:0.77) {77};
    \node[profile value, text=black, anchor=east, xshift=-3pt, yshift=2pt] at (-90:0.73) {73};
    \node[profile value, text=black, anchor=south east, xshift=-2pt, yshift=3pt] at (-150:0.94) {94};
    \node[profile value, text=black, anchor=north east, xshift=-2pt, yshift=-3pt] at (150:0.882) {88};

    \draw[very thick, blue!70!black, fill=blue!70!black, fill opacity=0.035]
      (90:0.49) -- (30:0.89) -- (-30:0.85) -- (-90:0.72) -- (-150:0.942) -- (150:0.852) -- cycle;
    \foreach \a/\r in {90/0.49,30/0.89,-30/0.85,-90/0.72,-150/0.942,150/0.852} {
      \fill[blue!70!black] (\a:\r) circle (0.018);
    }
    \node[profile value, text=blue!70!black, anchor=north west, xshift=1pt, yshift=-2pt] at (90:0.49) {49};
    \node[profile value, text=blue!70!black, anchor=south east, xshift=-2pt, yshift=5pt] at (30:0.89) {89};
    \node[profile value, text=blue!70!black, anchor=north west, xshift=3pt, yshift=-4pt] at (-30:0.85) {85};
    \node[profile value, text=blue!70!black, anchor=west, xshift=3pt, yshift=4pt] at (-90:0.72) {72};
    \node[profile value, text=blue!70!black, anchor=north east, xshift=-2pt, yshift=-4pt] at (-150:0.942) {94};
    \node[profile value, text=blue!70!black, anchor=south east, xshift=-2pt, yshift=4pt] at (150:0.852) {85};

    \draw[very thick, orange!85!black, fill=orange!85!black, fill opacity=0.035]
      (90:0.54) -- (30:0.95) -- (-30:0.74) -- (-90:0.72) -- (-150:0.764) -- (150:0.736) -- cycle;
    \foreach \a/\r in {90/0.54,30/0.95,-30/0.74,-90/0.72,-150/0.764,150/0.736} {
      \fill[orange!85!black] (\a:\r) circle (0.018);
    }
    \node[profile value, text=orange!85!black, anchor=south west, xshift=1pt, yshift=2pt] at (90:0.54) {54};
    \node[profile value, text=orange!85!black, anchor=south west, xshift=4pt, yshift=6pt] at (30:0.95) {95};
    \node[profile value, text=orange!85!black, anchor=north east, xshift=-2pt, yshift=-4pt] at (-30:0.74) {74};
    \node[profile value, text=orange!85!black, anchor=west, xshift=3pt, yshift=-5pt] at (-90:0.72) {72};
    \node[profile value, text=orange!85!black, anchor=south west, xshift=3pt, yshift=4pt] at (-150:0.764) {76};
    \node[profile value, text=orange!85!black, anchor=north east, xshift=-2pt, yshift=-3pt] at (150:0.736) {74};

    \draw[very thick, violet!80!black, fill=violet!80!black, fill opacity=0.035]
      (90:0.30) -- (30:0.93) -- (-30:0.62) -- (-90:0.37) -- (-150:0.748) -- (150:0.672) -- cycle;
    \foreach \a/\r in {90/0.30,30/0.93,-30/0.62,-90/0.37,-150/0.748,150/0.672} {
      \fill[violet!80!black] (\a:\r) circle (0.018);
    }
    \node[profile value, text=violet!80!black, anchor=north east, xshift=-1pt, yshift=-2pt] at (90:0.30) {30};
    \node[profile value, text=violet!80!black, anchor=north west, xshift=4pt, yshift=-6pt] at (30:0.93) {93};
    \node[profile value, text=violet!80!black, anchor=south east, xshift=-2pt, yshift=3pt] at (-30:0.62) {62};
    \node[profile value, text=violet!80!black, anchor=east, xshift=-3pt] at (-90:0.37) {37};
    \node[profile value, text=violet!80!black, anchor=north west, xshift=3pt, yshift=-4pt] at (-150:0.748) {75};
    \node[profile value, text=violet!80!black, anchor=south west, xshift=2pt, yshift=3pt] at (150:0.672) {67};

    \begin{scope}[shift={(1.33,0.24)}]
      \draw[very thick, black] (0,0) -- (0.19,0); \node[anchor=west] at (0.24,0) {Qwen3.6 35B};
      \draw[very thick, blue!70!black] (0,-0.14) -- (0.19,-0.14); \node[anchor=west] at (0.24,-0.14) {Qwen3.5 35B};
      \draw[very thick, orange!85!black] (0,-0.28) -- (0.19,-0.28); \node[anchor=west] at (0.24,-0.28) {GPT-OSS 20B};
      \draw[very thick, violet!80!black] (0,-0.42) -- (0.19,-0.42); \node[anchor=west] at (0.24,-0.42) {Qwen3 30B};
    \end{scope}
  \end{tikzpicture}
  }
  \caption{Verification profiles for four representative agents. Labels report percentages, and all axes run from $0$ to $100\%$ with higher better. \emph{Within budget} is one minus turn-budget exhaustion. \emph{Claim grounding} is one minus the largest hallucination-channel rate. \emph{Customer experience} is one minus mean trial-level maximum user frustration.}
  \label{fig:verification-profiles}
\end{figure}

The retained evidence distinguishes failures that a transcript review or final-cart rate would merge. Table~\ref{tab:diagnostics} reports the diagnostic channels, and Figure~\ref{fig:verification-profiles} combines six of them for four representative agents.

\paragraph{Target-cart evidence separates discovery from over-purchase.} Qwen3.5-35B reaches the highest recall in the sweep ($0.85$) and the highest goal-achievement rate ($0.66$), but its reconstructed final cart has precision $0.89$, below five agents it beats on goal achievement. It searched and added aggressively, issuing $44.1$ tool calls per conversation against a field median near $25$. The stored target cart and state replay identify the shopping failure. The assistant found items the customer wanted, then added items the target cart did not contain. A final success rate would not distinguish broad discovery from over-purchase.

\paragraph{Catalog evidence separates unsupported descriptions from wrong arithmetic.} GPT-OSS and Qwen3-30B fabricate product attributes at roughly four times the rate of the other six ($0.236$ and $0.252$ against a cluster near $0.055$), while their numeric claims, checked against catalog prices and reconstructed cart state, stay ordinary ($0.033$ and $0.029$). The catalog and tool trace localize the problem to unsupported descriptive claims rather than price, quantity, or total calculations. Qwen3-30B also has the highest phantom-action rate ($0.158$), announcing actions with no matching tool call.

\paragraph{Tool traces separate low activity from efficient shopping.} Qwen3-30B issues the fewest calls in the sweep at $15.4$ and is also the lowest exact-cart agent. Its transcripts show roughly one call per customer turn, and $63\%$ of trials exhaust the turn budget without the actions needed to close the cart. The tool trace and termination record show that its low call count reflects under-action rather than economy. The verifier also identifies premature cart writes, including GPT-OSS additions made before customer confirmation on $15.3\%$ of additions. Search traces add a separate relevance check. Catalog search returned candidates for every query in the sweep, so the bad-search rates in Table~\ref{tab:diagnostics} measure relevance rather than empty retrievals.

\paragraph{Termination records distinguish pace failures.} Gemma4-26B reaches exact-cart success on $0.27$ of light-exploration trials and $0.53$ of complex trials. It exhausts the budget on $22$ of $33$ light-exploration trials while issuing a median $1.14$ calls per turn, but exhausts only $1$ of $17$ complex trials at $2.04$ calls per turn. Qwen3.5-35B shows the opposite failure on complex baskets, issuing a median $66$ calls while $4$ of $17$ trials end in abandonment. These records distinguish agents that must move faster from agents that must use tools more selectively. Appendix~\ref{app:pace} gives the full analysis.

\begin{table}[t]
  \caption{Diagnostic channels. Lower is better for the failure-rate columns; call volume is descriptive. The three hallucination channels come from one metric split by claim type. Action and numeric claims are checked deterministically; attribute claims are checked by a judge.}
  \label{tab:diagnostics}
  \centering
  \small
  \setlength{\tabcolsep}{4pt}
  \begin{tabularx}{\textwidth}{@{}Yccccc@{}}
    \toprule
    & \multicolumn{3}{c}{Hallucination channels} & \multicolumn{2}{c}{Tool use} \\
    \cmidrule(r){2-4}\cmidrule(r){5-6}
    Agent & Phantom & Numeric & Attribute & Calls/conv & Bad search \\
    \midrule
    GPT-OSS 20B & 0.060 & 0.033 & 0.236 & 24.6 & 0.325 \\
    Gemma4 26B  & 0.081 & 0.045 & 0.063 & 21.4 & 0.315 \\
    Gemma4 31B  & 0.076 & \textbf{0.011} & 0.056 & 19.6 & \textbf{0.291} \\
    Qwen3 30B   & 0.158 & 0.029 & 0.252 & 15.4 & 0.365 \\
    Qwen3.5 35B & 0.055 & 0.018 & 0.058 & 44.1 & 0.415 \\
    Qwen3.5 27B & 0.084 & 0.019 & 0.060 & 30.4 & 0.434 \\
    Qwen3.6 35B & 0.060 & 0.027 & 0.055 & 32.8 & 0.404 \\
    Qwen3.6 27B & \textbf{0.050} & 0.042 & \textbf{0.050} & 26.1 & 0.352 \\
    \bottomrule
  \end{tabularx}
\end{table}

\paragraph{Cross-judge check.} We rescore $300$ matched shopping and grocery trials from Qwen3-30B and Qwen3.6-27B with GPT-5.4. Across five completed rubrics, the judges agree most closely on tool-call correctness (mean difference $0.002$, MAE $0.030$, $r=0.924$). Clarification has the largest MAE ($0.111$) and lowest correlation ($r=0.490$), while GPT-5.4 scores hallucination $0.052$ lower on average. This check covers two agents and does not calibrate either judge against human labels. Appendix~\ref{app:judge-agreement} reports the full protocol and results.

\section{Limitations}
\label{sec:limitations}

The main eight-agent sweep uses one primary judge. The two-agent check in Appendix~\ref{app:judge-agreement} covers five rubrics and excludes three whose secondary outputs were incomplete; neither judge has been calibrated against human labels. The attribute-hallucination and conversational scores therefore remain evaluator-dependent rather than externally validated customer judgments. Tool-call correctness also depends on the simulator's expected-call prediction. Errors in that prediction can change the score even when the assistant's tool trace is fixed.

Environment-grounded verification requires the target cart and reveal schedule to be fixed before a conversation begins. That structure lets the evaluator check cart outcomes and date actions against customer requests, but it cannot reproduce a shopper whose preferences remain unsettled, change during the interaction, or allow several acceptable products. The schedule fixes the scoring boundary; it does not make the simulator a human proxy. The user simulator also knows the hidden target and is directed to continue toward it. After an assistant fails to retrieve an item, we observe that the simulator can disclose an exact product name or another highly discriminating constraint that makes a further search easy. It can therefore rescue a poor search and act more cooperatively than a customer on live traffic. In a $\tau$-bench human comparison, general-purpose simulators were more cooperative and produced higher agent success than human users in most settings \citep{zhou2026sim2real}. With a GPT-4o agent fixed across $\tau$-bench retail tasks, changing the user model shifted success by nearly nine percentage points \citep{seshadri2026lost}.

The turn budget is a benchmark convention rather than an estimate of customer patience. It supports comparisons under one stopping policy, but another budget or tool-mediation policy may change the measured profiles. The checkpoint repository identifiers are recorded, while immutable repository revisions are not; exact weight provenance is therefore limited to the identifiers reported in Appendix~\ref{app:latency}. The synthetic catalog and standardized tool surface leave out conditions that shape live retail interactions. Prices and inventory do not change, and the environment does not include merchant-specific policies, delivery and returns, competing listings, or browser and interface friction. The results therefore characterize controlled, tool-mediated catalog shopping rather than performance on live retail traffic.

\section{Conclusion}

Final-cart success is insufficient for evaluating e-commerce agents because it does not identify the route an agent took to reach, miss, or exceed a customer's intended cart. We propose an e-commerce evaluation environment that precommits the target cart and reveal schedule against a fixed control catalog, then retains the tool trace and environment state produced during the trial. The verifier uses this evidence to establish cart correctness, search quality, action timing, and numeric catalog claims. Model-based judges assess attribute support and other questions that require language.

Across eight open-weight agents, we use the resulting capability profiles to identify distinct repair targets in tool use, cart management, catalog grounding, and retrieval. E-commerce evaluations should report these distinctions when deployment and repair decisions depend on how an agent fails. This controlled, tool-mediated design provides a reproducible pre-deployment test and complements validation on live retail traffic.

The broader lesson is that reliable agent verification requires environments to preserve the evidence needed to evaluate intermediate actions against the state and information available when those actions were taken. Environment-grounded verification can therefore turn terminal outcomes into actionable diagnoses and provide a general framework for evaluating how agents behave, not only whether they succeed.

\clearpage

\bibliographystyle{plainnat}
\bibliography{refs}

\appendix

\section{Per-retailer catalog composition}
\label{app:catalog}

Table~\ref{tab:app-catalog} gives the within-store share of each canonical category for the six retailers aggregated in Section~\ref{sec:env}. We derive the Amazon shares from published category counts for products observed in Amazon Reviews 2023 records from May 1996 through September 2023 \citep{hou2024bridging}. We exclude its \texttt{Unknown} bucket, which contains approximately $13.2$ million of $48.19$ million items, and renormalize the $33$ named source categories to $100\%$. The other five retailer shares derive from live department listing counts scraped from public storefronts. We map each retailer's taxonomy to the shared categories and assign a $0\%$ share when a retailer does not carry a category. Categories carried by fewer than three retailers are folded into \emph{Other}. We then average the six within-retailer shares for each shared category and normalize the aggregate vector to $100\%$. The analysis retains aggregate department counts and normalized shares; it does not redistribute reviews, product records, or storefront pages.

\begin{table}[h]
  \caption{Within-store category share (\%) for each retailer and the aggregate used to populate the control catalog.}
  \label{tab:app-catalog}
  \centering
  \small
  \setlength{\tabcolsep}{3pt}
  \begin{tabularx}{\textwidth}{@{}Yrrrrrrr@{}}
    \toprule
    Category & Amazon & Costco & Target & Walmart & Flipkart & M.\,Libre & Aggregate \\
    \midrule
    Clothing \& Accessories  & 22.99 & 15.67 & 24.33 & 5.99 & 11.23 & 11.24 & 15.24 \\
    Home \& Kitchen          & 10.60 & 11.46 & 23.35 & 20.30$^{\ast}$ & 9.14 & 15.00 & 14.97 \\
    Books \& Media           & 21.55 & 0.00 & 25.05 & 19.47 & 4.98 & 17.92 & 14.83 \\
    Electronics \& Computers & 8.56 & 7.56 & 2.51 & 4.64 & 46.96 & 7.45 & 12.95 \\
    Grocery \& Food          & 1.73 & 18.92 & 1.81 & 3.76 & 1.05 & 2.04 & 4.88 \\
    Sports \& Outdoors       & 4.58 & 4.60 & 4.24 & 4.59 & 5.11 & 4.21 & 4.56 \\
    Automotive               & 5.73 & 0.00 & 0.00 & 7.45 & 2.34 & 11.13 & 4.44 \\
    Home Improvement \& Tools& 4.30 & 5.30 & 0.00 & 4.13 & 1.90 & 6.80 & 3.74 \\
    Health \& Personal Care  & 2.46 & 7.99 & 2.77 & 3.89 & 2.95 & 1.02 & 3.51 \\
    Patio, Lawn \& Garden    & 2.44 & 8.09 & 3.50 & 4.94 & 0.00 & 1.31 & 3.38 \\
    Toys \& Games            & 2.94 & 0.00 & 5.08 & 4.41 & 2.12 & 4.38 & 3.16 \\
    Arts, Crafts \& Party    & 2.29 & 0.00 & 1.50 & 7.87 & 1.80 & 4.81 & 3.05 \\
    Appliances               & 0.27 & 12.87 & 0.00 & 0.00 & 2.95 & 1.76 & 2.97 \\
    Beauty                   & 3.19 & 2.99 & 2.05 & 1.50 & 2.52 & 3.26 & 2.58 \\
    Office \& School         & 2.03 & 0.00 & 1.60 & 2.89 & 2.52 & 0.00 & 1.51 \\
    Other                    & 1.70 & 1.78 & 0.00 & 0.00 & 0.00 & 4.04 & 1.25 \\
    Baby                     & 0.62 & 0.76 & 1.21 & 2.15 & 1.19 & 1.20 & 1.19 \\
    Pet Supplies             & 1.41 & 2.01 & 0.99 & 1.06 & 0.00 & 1.53 & 1.17 \\
    Musical Instruments      & 0.61 & 0.00 & 0.00 & 0.97 & 1.25 & 0.90 & 0.62 \\
    \bottomrule
  \end{tabularx}

  \vspace{3pt}
  {\footnotesize $^{\ast}$Walmart's Home \& Kitchen count excludes made-to-order listings. Digitally customizable frames account for a large share of that department, and counting them would distort the category mix of a general marketplace.}
\end{table}

The retailer columns in Table~\ref{tab:app-catalog} contain different category concentrations, while the aggregate distributes mass across the shared taxonomy. We average across these differences so no single retailer's catalog structure determines the control catalog.

\section{Basket construction parameters}
\label{app:quantity}

Scenario construction is deterministic from one integer seed. Reveal style, journey, waves, basket breadth and depth, quantities, and probe contents use separate keyed sub-generators. Persona traits and attribute dropout instead share one ordered generator.

\paragraph{Price weighting and depth.} A category's representative is drawn with weight $\exp\!\big(-(\log_{10}p - \mu)^2 / 2\sigma^2\big)$, where $\mu$ is the median log-price of the category's in-stock rows and $\sigma=0.6$. The spread is wide enough that the bulk of a category is sampled almost flatly and only the far tails are suppressed, which reproduces mid-range items being bought about as often as one another while a flagship is rare. The remaining $D_j - 1$ items in the slot are filled by the same weighting, drawn without replacement. The geometric depth of Section~\ref{sec:env} is truncated at $12$ products from any one category.

\paragraph{Breadth slots and distinct SKUs.} The $B$ category slots are drawn with replacement, so a category can come up more than once and the basket runs deeper there than breadth alone would give. A product-id exclusion set spans the whole basket. Each slot considers only in-stock rows not already taken, and each drawn product is added to the set before the next slot. A repeated category therefore runs a fresh representative and fill pass over what is left rather than duplicating a line, and every goal item names a distinct SKU. A slot whose category has no rows left contributes nothing, which is why the turn budget of Section~\ref{sec:env} is computed over the realized product count $m$ rather than the drawn breadth.

\paragraph{Quantities.} Each selected product receives a quantity $q\in[1,Q_{\max}]$ drawn from
\[
w(q)\;\propto\;(q+c)^{-\alpha}\,\cdot\,\underbrace{\exp\!\big(-\max(q-k,\,0)/\tau\big)}_{\text{grocery taper only}}.
\]
Non-grocery items use $(c,\alpha)=(0,2.7)$ with no taper, so $w(q)\propto q^{-2.7}$ falls steeply from $q=1$, with $w(1)/w(2)=2^{2.7}\approx 6.5$. The shopper almost always takes one. Grocery staples use $(c,\alpha)=(4,2.5)$, where the offset keeps $(q+c)$ nearly constant across small $q$ and flattens the head, giving $w(1)/w(2)=(6/5)^{2.5}\approx 1.6$ so quantities one through five are comparably likely. The taper past $k=5$ with $\tau=1$ suppresses implausible party-size counts.

\section{Persona dimensions}
\label{app:persona}

The simulated customer is drawn along six independent categorical dimensions (Table~\ref{tab:app-persona}), each sampled from the trial seed with realistic-frequency weights, so \texttt{moderate} specificity and \texttt{conversational} style are common while \texttt{garbled} input is rare. The sampled traits are written into the persona prompt the simulator follows.

\begin{table}[h]
  \caption{The six persona dimensions.}
  \label{tab:app-persona}
  \centering
  \small
  \begin{tabularx}{\textwidth}{l p{0.30\textwidth} Y}
    \toprule
    Dimension & Values & What it varies \\
    \midrule
    \texttt{specificity}      & none, vague, moderate, precise & How precisely the customer states what they want. \\
    \texttt{decisiveness}     & quick\_decider, comparison\_shopper, indecisive, impulsive & How readily they commit rather than compare or change their mind. \\
    \texttt{price\_sensitivity}& budget\_conscious, value\_seeker, price\_indifferent, luxury\_expecting & How strongly price drives the decision. \\
    \texttt{knowledge}        & novice, informed, expert & Domain expertise and product vocabulary. \\
    \texttt{communication}    & terse, conversational, verbose & Message length and style. \\
    \texttt{input\_quality}   & clean, sloppy, garbled & Typos, casing, and messiness of phrasing. \\
    \bottomrule
  \end{tabularx}
\end{table}

\section{Reveal-style sampling and journey construction}
\label{app:reveal}

A basket of fewer than two products has nothing to sequence and is trivially \texttt{batch}, except for a complex non-grocery trial that drew a single item, which falls back to \texttt{linear}. Everything larger draws a style from Table~\ref{tab:app-reveal}, keyed on the realized basket size and the trial kind rather than the tier label alone, since the category-first picker can hand a nominally light trial several products.

\begin{table}[h]
  \caption{Reveal-style draw weights for baskets of two or more products. Grocery never draws \texttt{nonlinear}; the invariant is enforced twice, once in the weights and once as a guard.}
  \label{tab:app-reveal}
  \centering
  \small
  \begin{tabularx}{\textwidth}{l Y}
    \toprule
    Trial kind ($\geq 2$ products) & Style weights \\
    \midrule
    Grocery, any tier                   & \texttt{batch} $0.70$ / \texttt{linear} $0.30$ \\
    Non-grocery complex                 & \texttt{linear} $0.40$ / \texttt{nonlinear} $0.60$ \\
    Other multi-item non-grocery        & \texttt{batch} $0.50$ / \texttt{linear} $0.50$ \\
    \bottomrule
  \end{tabularx}
\end{table}

\paragraph{Micro-batching.} A \texttt{batch} trial carrying more than four products is partitioned once at build time into ordered waves. The first wave holds two items, and each later wave targets a length drawn uniformly in $[2,5]$. Outside the complex tier the partition respects category contiguity, packing whole categories toward each wave's target and cutting a category larger than the maximum wave into random pieces; complex trials shuffle the items first. The controller releases the next wave only once every item in the current one is in the cart, so the revealed set grows a wave per turn rather than all at once.

\paragraph{Nonlinear beat construction.} Each goal item receives one essential add beat. When optional-beat budget remains, the generator inserts an introduction and may add an exploration beat. It defers each nonfinal item with probability $0.5$ while optional-beat budget remains. Before introducing a subsequent item, it may select one pending item uniformly for an early revisit; after processing the list, it returns to any items still pending in shuffled order. The resulting itinerary can therefore interleave unresolved items without following either basket order or a first-in, first-out return order.

\paragraph{Per-turn tool rounds.} The ceiling on tool-call rounds within a single agent turn is $10$ for the sequential styles, which trickle roughly one item per turn. A batch reveal can land a whole wave at once, so it earns two further rounds, one search and one add, for every item that can arrive together, capped by the largest wave. A trial with a large micro-batched basket therefore allows up to $20$ rounds in a turn. The generous ceiling stops a runaway turn from pinning a worker, and a turn that hits it is recorded as an agent error.

\paragraph{Beat compression.} A nonlinear itinerary is compiled to fit the turn budget, with $n$ essential add beats always emitted and flavor beats consuming only the $\max(0, \text{max\_turns} - n)$ turns left over. Because a trial can still fall behind at run time, the controller drops flavor beats from the front of the remaining itinerary whenever the turns left no longer exceed the essential adds whose items are still missing from the cart, so the basket stays completable under any pace the agent sets.

\subsection{Recorded reveal journeys}
\label{app:reveal-examples}

The following three examples retain the recorded customer and assistant wording, with \(\ldots\) marking omitted text or turns. The controller recorded exact-cart success and goal completion for all three interactions.

\paragraph{Example 1: nonlinear deferral, pivot, and return.} The customer in the Gemma4-31B interaction moves among a thermal tee, trousers, and loungewear before closing all three. The excerpt shows the controller maintaining unresolved items across several natural deferrals and returns.

\begin{quote}
\footnotesize
\textbf{Customer:} ``so yeah the thermel v-neck thing is definately top priority but actualy now that im thinkin about it i might also need some trousers too \ldots anyway lets start with the long sleev tee first tho''\\
\textbf{Customer:} ``hmm yeah those both look pretty solid actualy, lemme think on those for a sec tho. so in the meantime i was also thinkin i need some trousers like i mentioned earlier \ldots''\\
\textbf{Customer:} ``but hold on lemme ask you somthing about the thermel v-neck tees you showed me earlier \ldots also just so i dont forget later i probly need some loungewear too \ldots''\\
\emph{[five turns omitted]}\\
\textbf{Customer:} ``but actualy hold on, i just remembered i was still thinkin about those trousers earlier \ldots so yeah lets go with those trousers, and also that terry cloth lounge set \ldots''\\
\textbf{Customer:} ``and yeah i think im gonna grab that northfield basics thermel v-neck too actualy \ldots so yeah all three of those, lets get em in the cart''\\
\textbf{Assistant:} ``Done and done! I've added all three to your cart: the Grid Check trousers in Large, the Terry Cloth lounge set in Medium, and that midweight thermal v-neck in Medium.''
\end{quote}

\paragraph{Example 2: a large cart in micro-batches.} The reveal schedule for Qwen3.6-35B partitions seven target items into waves of $[2,2,2,1]$. The excerpt shows exact quantities and new categories arriving in bounded groups rather than as one opening list.

\begin{quote}
\footnotesize
\textbf{Customer:} ``I need unsweetened almond milk in the larger 1.89 L format, preferably non-GMO, and I'm also after an organic plain oat yogurt around 400 g. I've got a few other categories on my list too, but let's start with these.''\\
\textbf{Customer:} ``Actually, I'd like to go with the Meadowblend almond milk instead of the Dawnfield \ldots Can you add five units of the Meadowblend almond milk and three of the Rootwell oat yogurt to my cart?''\\
\textbf{Customer:} ``Great! Now I also need one unit of unsweetened coconut milk, preferably organic in a 400 mL format, and two units of cashew-style aged vegan cheese around 150 g \ldots''\\
\textbf{Customer:} ``Perfect! Next on my list: I need six units of a photographic cityscape framed print, preferably in monochrome with a black metal frame around 60x90 cm. And I also need one unit of a brightening vitamin C eye cream for normal skin.''\\
\textbf{Customer:} ``Last thing on my list today: I need one unit of a modular desk organizer in navy leatherette. Something that can keep my workspace tidy and looks professional.''\\
\textbf{Assistant:} ``Perfect! I've added the last item to your cart: \ldots Your complete cart now includes: \ldots You're all set!''
\end{quote}

\paragraph{Example 3: one completed choice at a time.} The reveal schedule for Qwen3.6-35B introduces a flannel, a polo, and pruning shears in sequence. Each new request follows the preceding choice, providing a direct contrast with wave-based and nonlinear disclosure.

\begin{quote}
\footnotesize
\textbf{Customer:} ``I'm looking for a men's relaxed-fit flannel shirt with a button-down collar, ideally in a gingham check pattern or something similar \ldots''\\
\textbf{Customer:} ``Perfect, that's exactly what I need \ldots Yes, please add that to my cart. Now, I'm also looking for a men's slim-fit polo shirt with a spread collar \ldots''\\
\textbf{Customer:} ``The Coastal Breeze at \$28.33 is perfect \ldots Please add that one to my cart in Medium. Now, I also need to grab a pair of forged-steel pruning shears for my garden \ldots''\\
\textbf{Customer:} ``Yes, absolutely \ldots Please add those to my cart as well. That should do it for me today \ldots can you give me a quick rundown of what's in my cart and the total?''\\
\textbf{Assistant:} ``Perfect! Here's everything in your cart: \ldots Alpine Flannel Relaxed Weekend Shirt \ldots Coastal Breeze Spread-Collar Polo \ldots CrispCut Forged Pruning Shears \ldots''
\end{quote}

\section{Tool categories and extra-call penalties}
\label{app:toolcat}

Each tool's category is a fixed property read from the tool registry, which is how the tool-call-correctness metric decides whether a call mutates the cart. The $k$-th unjustified extra past a category's free allowance costs $\text{initial} + (k-1)\cdot\text{increase}$, so per-extra cost grows linearly and the cumulative penalty grows quadratically. We report the registry mapping and penalty schedule in Table~\ref{tab:app-toolcat}. An unmatched cart mutation incurs the full penalty from its first occurrence, while read-only and exploratory calls receive limited allowances.

\begin{table}[h]
  \caption{Extra-call penalty by tool category.}
  \label{tab:app-toolcat}
  \centering
  \small
  \setlength{\tabcolsep}{3pt}
  \begin{tabularx}{\textwidth}{@{}Y Y c Y@{}}
    \toprule
    Category & Tools & Free / turn & Cost of each extra beyond the allowance \\
    \midrule
    \path{READ_ONLY}         & \path{view_cart}, \path{get_product_details}, \path{check_availability} & 3 & $0.1,\ 0.2,\ 0.3,\ \dots$ \\
    \path{WRITE_EXPLORATORY} & \path{apply_coupon} & 2 & $0.25,\ 0.5,\ 0.75,\ \dots$ \\
    \path{EXPLORE}           & \path{search_catalog}, \path{get_recommendations} & 1 & $0.5,\ 0.75,\ 1.0,\ \dots$ \\
    \path{WRITE}             & \path{add_to_cart}, \path{remove_from_cart} & 0 & $1.0$ each \\
    \bottomrule
  \end{tabularx}
\end{table}

\section{Verification case studies}
\label{app:cases}

The following three examples show how the verifier connects a short customer exchange to the evidence that supports each result. We retain product identifiers because the tool log and catalog rows use them as stable keys.

\paragraph{Example 4: exact cart with an unsupported description.} The customer first asked GPT-OSS for caramel candies and then requested black licorice. The assistant recommended ``Twisted, slightly crunchy twists with a hint of anise and fennel'' for the product \texttt{GC031}. The catalog description for that product states that the twists are chewy and does not support the crunchy-texture claim. The customer instead selected \texttt{GC010}; state replay recovered the requested final cart exactly: three \texttt{GC012}, two \texttt{GC009}, and four \texttt{GC010}. The trial therefore has exact-cart success $1.0$ and zero numeric or phantom-action violations, while the attribute checker records an attribute-hallucination rate of $1.0$. A final-cart score reports success; the catalog-grounded claim check identifies a separate repair target.

\begin{quote}
\small
\textbf{Customer:} ``One more thing. I'd also like to get some black licorice. Do you carry any of those?''\\
\textbf{Assistant:} ``Amber Canopy Black Licorice Twists \ldots Twisted, slightly crunchy twists with a hint of anise and fennel.''\\
\textbf{Verifier:} the catalog row describes the twists as chewy; it contains no support for ``slightly crunchy.''
\end{quote}

\paragraph{Example 5: claimed cart action without a tool action.} The customer asked Qwen3-30B to add a Ridgeback trucker cap and a merino beanie. The assistant replied that both had been added. The following customer message reported that the cart did not change. Before the assistant's completion claim, the tool trace contains no matching cart-addition call. Later cart-addition calls occur, but they cannot support the earlier claim that both additions were already complete. The verifier therefore labels both action claims as phantom. The trial ends at the turn budget with recall $0.25$, cart precision $0.50$, phantom-action rate $1.0$, and attribute-hallucination rate $0.75$. This distinction requires the tool trace and state replay; the assistant's confirmation alone is insufficient evidence that the cart had changed at that point.

\begin{quote}
\small
\textbf{Customer:} ``Can you just add them both in?''\\
\textbf{Assistant:} ``Ridgeback Performance Trucker \ldots added. Merino Wool Ribbed Ski Beanie \ldots added.''\\
\textbf{Tool record:} no matching cart-mutation call precedes the completion claim; later cart calls do occur.
\end{quote}

\paragraph{Example 6: goal achievement with substantial over-purchase.} The customer opened the Qwen3.5-35B interaction with ``hey need some shirts n sneakers'' and later selected individual shirt and sneaker options. The final cart satisfied every target requirement, so goal achievement was true. State replay nevertheless found $28$ distinct products in the cart for a $15$-SKU target, including $9$ units of \texttt{SN453} where the target requested one. The final cart had recall $1.0$ but precision $0.344$, so exact-cart success was false. The case distinguishes discovering all wanted products from managing the cart conservatively.

\section{Verification cost}
\label{app:cost}

We separate total token volume from the metered cost of post-hoc judging in Tables~\ref{tab:app-tokens} and~\ref{tab:app-rubric-cost}. The agent under test accounts for more tokens per trial than the user simulator and judge combined, although those tokens come from the self-hosted inference path.

Across the eight-model sweep, the assistant averages $417$K tokens per trial, while the user simulator and judge add $347$K. Metered simulator and judge calls cost \$94 per $160$-trial run on average (\$0.59 per trial).

\begin{table}[h]
  \caption{Mean tokens per trial (thousands), averaged over the eight-model sweep.}
  \label{tab:app-tokens}
  \centering
  \small
  \begin{tabular}{lrrr}
    \toprule
    Party & Input & Output & Total \\
    \midrule
    Agent under test (self-hosted) & 410.9 & 5.7 & 416.6 \\
    User simulator (metered)       & 253.8 & 3.2 & 257.0 \\
    Judge (metered)                & 79.0 & 11.2 & 90.2 \\
    \bottomrule
  \end{tabular}
\end{table}

Median per-trial totals run to about $146$K tokens for a general shopping trial and $135$K for a grocery basket, but only $40$K for a refusal probe. The per-trial mean of $417$K sits well above these medians because a heavy tail of many-round baskets pulls it up.

Hallucination and tool-call correctness are the two largest judge costs in Table~\ref{tab:app-rubric-cost}. Both operate on long, tool-rich conversations, while the refusal rubric runs on only ten probes per model.

\begin{table}[h]
  \caption{Per-rubric judge cost, median over the eight-model sweep. \emph{Tokens/call} is median input plus output per judge call; \emph{Out.\ share} is the generated fraction, which costs more per token; \emph{USD/run} is the median cost across a $160$-trial run.}
  \label{tab:app-rubric-cost}
  \centering
  \small
  \begin{tabular}{lrrr}
    \toprule
    Rubric & Tokens/call & Out.\ share & USD/run \\
    \midrule
    Hallucination            & 23.1\,K & 23\% & 17.79 \\
    Tool-call correctness    & 19.6\,K & 10\% & 11.65 \\
    Add-to-cart behavior     & 9.6\,K  & 17\% & 6.72 \\
    Clarification            & 11.5\,K & 10\% & 5.68 \\
    Assistant coherence      & 8.5\,K  & 7\%  & 4.96 \\
    Recommendation usefulness& 8.4\,K  & 6\%  & 4.62 \\
    Intent understanding     & 8.2\,K  & 5\%  & 4.42 \\
    Response helpfulness     & 7.6\,K  & 12\% & 3.67 \\
    Refusal resistance$^{\ast}$ & 7.5\,K & 16\% & 0.26 \\
    \bottomrule
  \end{tabular}

  \vspace{3pt}
  {\footnotesize $^{\ast}$Cheap only because it runs over a $10$-probe cohort rather than the full sweep; its per-call footprint matches the other whole-conversation rubrics.}
\end{table}

\section{Model identity and latency}
\label{app:latency}

The sweep uses the base checkpoints \path{openai/gpt-oss-20b}, \path{google/gemma-4-26B-A4B-it}, \path{google/gemma-4-31B-it}, \path{Qwen/Qwen3-30B-A3B}, \path{Qwen/Qwen3.5-35B-A3B}, \path{Qwen/Qwen3.5-27B}, \path{Qwen/Qwen3.6-35B-A3B}, and \path{Qwen/Qwen3.6-27B}. Official reports and cards document these model families and their architectures \citep{openai2025gptoss,yang2025qwen3,qwen2026qwen35,qwen2026qwen3635,qwen2026qwen3627,gemma2026gemma4}.

We report both per-round-trip and per-trial latency in Table~\ref{tab:app-latency}. The five sparse-mixture agents occupy the five fastest per-round-trip rows, while the three dense agents require more time for each exchange. Per-trial latency also depends on how many exchanges an agent uses, so it does not follow the per-round-trip ordering exactly.

Each run uses tensor parallelism across two 80GB NVIDIA H100 GPUs and takes $1.8$--$10.4$ wall-clock hours (mean $3.3$), for $53.3$ H100-hours across the eight sweeps.

\begin{table}[h]
  \caption{Agent latency, ordered by per-round-trip time. \emph{Active} is the rounded number of parameters activated per token reported in first-party model documentation. Every sparse-mixture model is faster per round-trip than every dense model, so nominal size does not order latency.}
  \label{tab:app-latency}
  \centering
  \small
  \begin{tabular}{llrr}
    \toprule
    Model & Active & s/round-trip & s/trial \\
    \midrule
    GPT-OSS 20B  & $\approx$3.6B (MoE) & 0.6 & 19.7 \\
    Gemma4 26B   & 4B (MoE)   & 1.6 & 36.3 \\
    Qwen3.6 35B  & 3B (MoE)   & 2.0 & 58.5 \\
    Qwen3.5 35B  & 3B (MoE)   & 2.3 & 76.2 \\
    Qwen3 30B    & 3B (MoE)   & 2.3 & 49.3 \\
    Gemma4 31B   & 31B (dense)& 3.9 & 82.9 \\
    Qwen3.5 27B  & 27B (dense)& 5.7 & 156.3 \\
    Qwen3.6 27B  & 27B (dense)& 6.4 & 159.6 \\
    \bottomrule
  \end{tabular}
\end{table}

\section{The metric battery}
\label{app:metrics}

Each run reports the $44$ metrics in Table~\ref{tab:app-metrics}, every one of them a mean over the trials that produced a valid score. The harness stamps the scoring-contract version on each transcript and score artifact. Keys are the field names in the run summary files, so each row traces to the artifact it came from. Termination reasons are counted as they occur rather than declared in advance, so the five below are the reasons this sweep produced; the sixth the harness can record, a failed opening message, never fired. We group deterministic outputs, the user simulator's inline rating, and post-hoc judge scores separately. Tables~\ref{tab:app-outcome} and~\ref{tab:app-quality} report the subset that varies across agents.

\begingroup
  \small
  \setlength{\tabcolsep}{4pt}
  \begin{longtable}{@{}>{\raggedright\arraybackslash}p{0.34\textwidth}>{\raggedright\arraybackslash}p{0.62\textwidth}@{}}
    \caption{The $44$ metrics produced by a run. Rates are in $[0,1]$; rubric scores are normalized to $[0,1]$ with higher better.}
    \label{tab:app-metrics} \\
    \toprule
    Metric (\texttt{summary key}) & Definition \\
    \midrule
    \endfirsthead
    \multicolumn{2}{c}{Table~\ref{tab:app-metrics} continued} \\
    \toprule
    Metric (\texttt{summary key}) & Definition \\
    \midrule
    \endhead
    \midrule
    \multicolumn{2}{r}{Continued on next page} \\
    \endfoot
    \bottomrule
    \endlastfoot
    \multicolumn{2}{l}{\emph{Deterministic, replayed from the tool log (14)}} \\
    \midrule
    Goal achievement (\path{goal_rate}) & Fraction of trials where the cart holds at least the required quantity of every target product. \\
    Goal progress (\path{goal_progress}) & Highest fraction of exactly-satisfied target products reached at any point in the trial. \\
    Exact cart success (\path{exact_cart_success_rate}) & Cart holds exactly the target items at exactly the target quantities. \\
    Cart precision (\path{exact_cart_precision}) & Matched units over total units in the final cart. \\
    Cart recall (\path{exact_cart_recall}) & Matched units over total units in the target cart. \\
    Turns to goal (\path{avg_turns_to_goal}) & Customer turns before the goal was met, over trials that met it. \\
    Tool calls per conversation (\path{avg_total_tool_calls}) & Calls issued across the whole trial. \\
    Tool calls per successful conversation (\path{tool_calls_per_successful_conv}) & The same count restricted to trials that met the goal. \\
    Invalid-call rate (\path{invalid_tool_call_rate}) & Calls whose result reports an error, over all calls. \\
    Redundant-call rate (\path{redundant_tool_call_rate}) & Repeated call signatures within a turn with no intervening state change, over all calls. \\
    Empty-search rate (\path{empty_search_rate}) & Searches that executed cleanly and returned nothing, over all searches. \\
    Bad-search rate (\path{bad_search_rate}) & Goal-relevant searches returning none of the still-needed targets, over goal-relevant searches. \\
    Recovery, overall (\path{search_recovery_rate_overall}) & Bad searches followed by a successful same-category retry. \\
    Recovery, autonomous (\path{search_recovery_rate_autonomous}) & The same retry within the same turn, with no customer re-prompt. \\
    \midrule
    \multicolumn{2}{l}{\emph{Exact-cart success split by reporting tier (3)}} \\
    \midrule
    Simple (\path{difficulty.simple_success}) & Direct purchase and light exploration, difficulties $1$--$5$. \\
    Multi-item (\path{difficulty.multi_item_success}) & Difficulties $6$--$8$. \\
    Complex (\path{difficulty.complex_success}) & Difficulties $9$--$10$. \\
    \midrule
    \multicolumn{2}{l}{\emph{Rated inline by the user simulator (1)}} \\
    \midrule
    User frustration (\path{inline.user_frustration}) & Worst per-turn frustration the simulated customer reported in the trial. \\
    \midrule
    \multicolumn{2}{l}{\emph{How the trial ended (5)}} \\
    \midrule
    Goal completion (\path{goal_met}) & Every target item reached its required quantity. \\
    Turn budget exhausted (\path{max_turns}) & The basket-aware turn budget ran out first. \\
    Customer abandoned (\path{abandoned}) & The simulated customer gave up in character. \\
    Assistant error (\path{assistant_error}) & The agent under test failed to produce a turn. \\
    Simulator error (\path{judge_error}) & The user simulator failed to produce a turn. \\
    \midrule
    \multicolumn{2}{l}{\emph{Rubric scores, scored on the finished transcript (8)}} \\
    \midrule
    Hallucination (\path{hallucination}) & One minus the violation rate over all extracted claims, across the three channels below. \\
    Response helpfulness (\path{response_helpful}) & Whether each reply moved the purchase forward. \\
    Assistant coherence (\path{assistant_coherence}) & Whether the agent remembered what the customer had already said. \\
    Intent understanding (\path{intent_understanding}) & Whether the agent read the request correctly. \\
    Clarification (\path{clarification}) & Whether questions were asked when needed and skipped when not. Policy-aware. \\
    Recommendation usefulness (\path{recommendation_usefulness}) & Whether what was surfaced fits what the customer asked for. \\
    Add-to-cart behavior (\path{add_to_cart_behavior}) & Whether variant and quantity were confirmed before the add. \\
    Tool-call correctness (\path{tool_call_custom}) & Per-turn agreement with the simulator's predicted call chain, taxed for unjustified extras. \\
    \midrule
    \multicolumn{2}{l}{\emph{Typed failure rates emitted alongside the rubrics (6)}} \\
    \midrule
    Phantom action (\path{phantom_action_rate}) & Announced cart actions with no matching tool call. Checked against the tool log. \\
    Numeric hallucination (\path{numeric_hallucination_rate}) & Price, stock, or cart-total claims the catalog contradicts. Checked against catalog rows. \\
    Attribute hallucination (\path{attribute_hallucination_rate}) & Product-attribute claims not entailed by the catalog row. Judged. \\
    Premature add-to-cart (\path{false_add_to_cart}) & Adds made before the customer confirmed variant and quantity. \\
    Unnecessary clarification (\path{unnecessary_clarification_rate}) & Questions asking for what the customer had already given. \\
    Failure to clarify (\path{missed_clarification_rate}) & Missing questions the request required. \\
    \midrule
    \multicolumn{2}{l}{\emph{Refusal probes, $10$ per model (7)}} \\
    \midrule
    Refusal index (\path{censorship_resistance}) & Mean probe score, $1.0$ an ordinary shopping interaction and $0.0$ a hard refusal with a lecture. \\
    Hard refusal (\path{hard_refusal}) & The agent declined to shop for the requested merchandise. \\
    Moral framing (\path{political_or_moral_framing}) & The agent volunteered a position on the topic. \\
    Search attempted (\path{search_attempted}) & The agent searched the catalog rather than refusing outright. \\
    Faithful search terms (\path{search_terms_faithful_mean}) & How completely the query preserved the customer's politically-marked wording. \\
    Transparent empty result (\path{zero_result_handled_transparently}) & An empty result was reported honestly rather than papered over. \\
    Off-theme redirect (\path{unrelated_redirect_offered}) & The agent steered toward unrelated merchandise. \\
\end{longtable}
\endgroup

\section{Refusal probes}
\label{app:probes}

A separate cohort of ten probes per model tests whether an agent applies political or identity-motivated censorship to lawful shopping requests. Each probe asks for merchandise associated with Chinese political or LGBTQ topics and uses a synthetic placeholder target that the control catalog does not contain. The probe reuses the ordinary simulator and logging path but replaces cart scoring with a policy-aware rubric. The rubric receives the agent's system prompt, anchors an ordinary shopping interaction at $1.0$ and a spontaneous hard refusal with a lecture at $0.0$, and records six typed behavioral flags.

The runner consults the applicability map, applies this rubric only to refusal probes, and records the shopping rubrics as skipped rather than zero. It therefore does not score the unavailable target as a shopping failure. Table~\ref{tab:app-refusal} reports the resulting score and flags. The agents attempted a catalog search on $0.80$--$1.00$ of probes, while hard-refusal rates ranged from $0.00$ to $0.20$. The cohort supports only directional comparisons because it contains ten probes per model. Rates for search faithfulness and transparent handling can use smaller denominators because the rubric records them only when a search or an empty result occurs. On the $20$ matched probes from the two-agent cross-judge check, the refusal scores correlate at $r=0.952$, but GPT-5.4 scores them $0.100$ lower on average (MAE $0.120$). This offset also remains directional because the paired sample is small.

\begin{table}[h]
  \caption{Refusal-probe metrics over $10$ probes per model. Boolean rates with complete coverage change in increments of $0.10$. \emph{Faithful terms} measures how completely the agent preserved the customer's politically-marked wording in its search query; this flag and \emph{transparent empty result} exclude probes where the relevant action did not occur.}
  \label{tab:app-refusal}
  \centering
  \small
  \setlength{\tabcolsep}{3pt}
  \begin{tabularx}{\textwidth}{@{}Ycccccccc@{}}
    \toprule
    & GPT-OSS & Gem.\,26B & Gem.\,31B & Q3\,30B & Q3.5\,35B & Q3.5\,27B & Q3.6\,35B & Q3.6\,27B \\
    \midrule
    Refusal index $\uparrow$   & 0.86 & 0.85 & 0.87 & 0.69 & 0.69 & 0.84 & 0.88 & 0.78 \\
    Hard refusal               & 0.00 & 0.00 & 0.10 & 0.20 & 0.20 & 0.00 & 0.00 & 0.20 \\
    Moral framing              & 0.00 & 0.00 & 0.00 & 0.20 & 0.20 & 0.30 & 0.00 & 0.30 \\
    Search attempted           & 1.00 & 1.00 & 0.90 & 0.80 & 0.90 & 1.00 & 1.00 & 0.90 \\
    Off-theme redirect         & 0.60 & 0.40 & 0.20 & 0.40 & 0.50 & 0.40 & 0.60 & 0.30 \\
    Faithful terms $\uparrow$  & 1.00 & 0.90 & 0.99 & 1.00 & 0.87 & 0.87 & 0.99 & 0.89 \\
    Transparent empty result   & 1.00 & 0.90 & 1.00 & 0.88 & 1.00 & 1.00 & 1.00 & 1.00 \\
    \bottomrule
  \end{tabularx}
\end{table}

\section{Cross-judge agreement}
\label{app:judge-agreement}

We rescore the $150$ shopping and grocery trials for Qwen3-30B and Qwen3.6-27B with \path{openai/gpt-5.4}, holding each transcript, tool trace, and simulator record fixed. We queried the model through OpenRouter on August 13, 2026 UTC with temperature zero, a $16{,}000$-token completion cap, and evaluation module \path{v3.1.0}; Anthropic prompt caching did not apply to these calls. This produces $300$ paired observations for each fully covered standard-trial rubric. We retain the five rubrics for which the secondary judge returned completed rationales. We exclude coherence, intent understanding, and recommendation usefulness because the secondary outputs contained incomplete evaluation placeholders rather than judgments. Table~\ref{tab:judge-agreement} reports each judge's mean, the signed paired difference, the mean absolute paired difference (MAE), and the Pearson correlation across trials.

\begin{table}[h]
  \caption{Primary and secondary judge scores on matched standard trials from two agents. $\Delta$ is GPT-5.4 minus Claude Sonnet 4.5. MAE measures trial-level disagreement; $r$ measures whether the two scores vary together. Coverage falls below $300$ when either judge returned an error or no valid score.}
  \label{tab:judge-agreement}
  \centering
  \small
  \setlength{\tabcolsep}{4pt}
  \begin{tabularx}{\textwidth}{@{}Yrrrrrr@{}}
    \toprule
    Metric & $n$ & Claude & GPT-5.4 & $\Delta$ & MAE & $r$ \\
    \midrule
    Tool-call correctness & 300 & 0.740 & 0.742 & $+0.002$ & 0.030 & 0.924 \\
    Response helpfulness  & 298 & 0.944 & 0.917 & $-0.027$ & 0.043 & 0.813 \\
    Add-to-cart behavior  & 300 & 0.919 & 0.880 & $-0.039$ & 0.077 & 0.681 \\
    Hallucination         & 279 & 0.903 & 0.851 & $-0.052$ & 0.087 & 0.602 \\
    Clarification         & 300 & 0.899 & 0.876 & $-0.023$ & 0.111 & 0.490 \\
    \bottomrule
  \end{tabularx}
\end{table}

The judges agree most closely on tool-call correctness. Its means differ by $0.002$, its MAE is $0.030$, and the trial-level correlation is $0.924$. Response helpfulness also remains close: the judges differ by at most $0.1$ on $91.3\%$ of paired trials, although GPT-5.4 scores it $0.027$ lower on average. These rubrics either combine fixed checks with a narrow model decision or grade one reply at a time.

Clarification produces the largest standard-trial MAE ($0.111$) and the lowest correlation ($0.490$), while its signed difference is smaller ($-0.023$). The judges therefore disagree on which individual conversations contain unnecessary or missed questions more than their aggregate means suggest. Hallucination shows a more directional difference: GPT-5.4 scores it $0.052$ lower on average, with MAE $0.087$. Add-to-cart behavior lies between these cases. Within this two-agent subset, aggregate conclusions are least sensitive to judge choice for tool-call correctness and response helpfulness, while clarification and claim assessment require stronger calibration.

Our manual review of several high-disagreement trials found that GPT-5.4 applied stricter penalties to unsupported attribute claims and redirects in refusal probes, while Claude Sonnet 4.5 applied stricter penalties in some clarification and response-helpfulness cases. From this targeted review, we cannot establish a universal ordering of judge strictness or accuracy. We have not calibrated either judge against independent human labels.

\section{Difficulty-tier outcomes and shopping pace}
\label{app:pace}

Table~\ref{tab:outcome} reports exact-cart success overall and by generator tier. The tier samples are small, so their intervals are wide. We use these results to identify behavioral variation rather than rank agents within tiers.

\begin{table}[h]
  \caption{Exact-cart success with $95\%$ bootstrap confidence intervals, overall and split by the four generator tiers.}
  \label{tab:outcome}
  \centering
  \small
  \setlength{\tabcolsep}{3pt}
  \begin{tabularx}{\textwidth}{@{}l*{5}{>{\centering\arraybackslash}X}@{}}
    \toprule
    & Overall ($n{=}100$) & Direct ($n{=}17$) & Light ($n{=}33$) & Multi-item ($n{=}33$) & Complex ($n{=}17$) \\
    \midrule
    GPT-OSS 20B  & 0.54 [0.44, 0.64] & 0.59 [0.35, 0.82] & 0.42 [0.27, 0.58] & \textbf{0.67} [0.52, 0.82] & 0.47 [0.24, 0.71] \\
    Gemma4 26B   & 0.45 [0.35, 0.55] & 0.59 [0.35, 0.82] & 0.27 [0.12, 0.42] & 0.52 [0.33, 0.70] & \textbf{0.53} [0.29, 0.76] \\
    Gemma4 31B   & 0.51 [0.41, 0.61] & \textbf{0.71} [0.47, 0.88] & 0.52 [0.33, 0.70] & 0.45 [0.30, 0.64] & 0.41 [0.18, 0.65] \\
    Qwen3 30B    & 0.30 [0.21, 0.39] & 0.29 [0.12, 0.53] & 0.30 [0.15, 0.45] & 0.30 [0.15, 0.45] & 0.29 [0.06, 0.53] \\
    Qwen3.5 35B  & 0.49 [0.39, 0.59] & 0.59 [0.35, 0.82] & \textbf{0.55} [0.36, 0.70] & 0.48 [0.33, 0.64] & 0.29 [0.12, 0.53] \\
    Qwen3.5 27B  & 0.41 [0.31, 0.51] & 0.47 [0.24, 0.71] & 0.52 [0.33, 0.70] & 0.30 [0.15, 0.45] & 0.35 [0.12, 0.59] \\
    Qwen3.6 35B  & \textbf{0.57} [0.47, 0.66] & \textbf{0.71} [0.47, 0.88] & \textbf{0.55} [0.36, 0.70] & 0.64 [0.45, 0.79] & 0.35 [0.12, 0.59] \\
    Qwen3.6 27B  & 0.53 [0.43, 0.63] & 0.53 [0.29, 0.76] & 0.52 [0.33, 0.70] & 0.64 [0.48, 0.79] & 0.35 [0.12, 0.59] \\
    \bottomrule
  \end{tabularx}
\end{table}

The generator gives larger baskets and less specific requests proportionally more turns. The turn budget therefore measures whether an agent can match the required shopping pace as well as whether it can assemble the target cart.

Gemma4-26B illustrates this relation. On light exploration, $22$ conversations reach the turn budget and $11$ reach the goal; only $1$ complex trial reaches the budget. The model issues a median $1.14$ tool calls per turn on light exploration, compared with $2.04$ on complex baskets. It spends early turns on rapport and mandatory size or gender clarification, then reaches the customer's confirmation after the light-exploration budget has closed. The complex-tier budget scales to $2.25$ turns per item and gives the same conversational style room to complete a larger basket.

Qwen3.5-35B shows the opposite pattern on complex baskets. It issues a median $66$ and up to $239$ calls in one conversation, while $4$ of $17$ trials end in customer abandonment. Termination reasons and tool-call density therefore identify whether an agent should move more directly from request to cart action or reduce unnecessary tool use.

\section{Full metric tables}
\label{app:full}

We report the complete outcome and tool-use results for all eight agents in Table~\ref{tab:app-outcome}. Qwen3-30B combines the lowest exact-cart success with the fewest calls and the highest turn-budget-exhaustion rate, consistent with under-action. Qwen3.5-35B instead reaches the highest goal-achievement rate and cart recall while issuing the most calls and retaining lower cart precision, consistent with over-purchase.

\begin{table}[h]
  \caption{Outcome and tool-use metrics over $100$ shopping trials per model. Turns to goal averages only over trials that reached the goal.}
  \label{tab:app-outcome}
  \centering
  \small
  \setlength{\tabcolsep}{3pt}
  \begin{tabularx}{\textwidth}{@{}Ycccccccc@{}}
    \toprule
    Metric & GPT-OSS & Gem.\,26B & Gem.\,31B & Q3\,30B & Q3.5\,35B & Q3.5\,27B & Q3.6\,35B & Q3.6\,27B \\
    \midrule
    Goal achievement        & 0.59 & 0.60 & 0.61 & 0.33 & 0.66 & 0.50 & 0.64 & 0.60 \\
    Goal progress           & 0.73 & 0.75 & 0.76 & 0.60 & 0.83 & 0.69 & 0.77 & 0.78 \\
    Exact cart success      & 0.54 & 0.45 & 0.51 & 0.30 & 0.49 & 0.41 & 0.57 & 0.53 \\
    Simple success          & 0.48 & 0.38 & 0.58 & 0.30 & 0.56 & 0.50 & 0.60 & 0.52 \\
    Multi-item success      & 0.67 & 0.52 & 0.45 & 0.30 & 0.48 & 0.30 & 0.64 & 0.64 \\
    Complex success         & 0.47 & 0.53 & 0.41 & 0.29 & 0.29 & 0.35 & 0.35 & 0.35 \\
    Cart precision          & 0.95 & 0.94 & 0.97 & 0.93 & 0.89 & 0.87 & 0.87 & 0.90 \\
    Cart recall             & 0.74 & 0.74 & 0.77 & 0.62 & 0.85 & 0.69 & 0.77 & 0.79 \\
    Turns to goal           & 8.8 & 10.7 & 9.1 & 10.3 & 8.8 & 8.6 & 8.5 & 7.1 \\
    \midrule
    Tool calls / conv.      & 24.6 & 21.4 & 19.6 & 15.4 & 44.1 & 30.4 & 32.8 & 26.1 \\
    Calls / successful conv.& 25.1 & 25.1 & 21.0 & 16.0 & 38.5 & 26.5 & 30.2 & 21.9 \\
    Redundant-call rate     & 0.004 & 0.002 & 0.001 & 0.000 & 0.008 & 0.002 & 0.005 & 0.003 \\
    Invalid-call rate       & 0.004 & 0.000 & 0.000 & 0.000 & 0.000 & 0.000 & 0.000 & 0.000 \\
    Empty-search rate       & 0.000 & 0.000 & 0.000 & 0.000 & 0.000 & 0.000 & 0.000 & 0.000 \\
    Bad-search rate         & 0.325 & 0.315 & 0.291 & 0.365 & 0.415 & 0.434 & 0.404 & 0.352 \\
    Recovery, overall       & 0.53 & 0.61 & 0.59 & 0.49 & 0.47 & 0.45 & 0.55 & 0.48 \\
    Recovery, autonomous    & 0.19 & 0.29 & 0.24 & 0.19 & 0.20 & 0.23 & 0.39 & 0.24 \\
    \midrule
    Goal completion (termination) & 0.60 & 0.61 & 0.63 & 0.31 & 0.65 & 0.50 & 0.64 & 0.63 \\
    Turn budget exhausted   & 0.28 & 0.38 & 0.36 & 0.63 & 0.28 & 0.41 & 0.27 & 0.32 \\
    Customer abandoned      & 0.03 & 0.01 & 0.01 & 0.04 & 0.04 & 0.02 & 0.04 & 0.03 \\
    Assistant error          & 0.03 & 0.00 & 0.00 & 0.00 & 0.02 & 0.07 & 0.05 & 0.02 \\
    Simulator error          & 0.06 & 0.00 & 0.00 & 0.02 & 0.01 & 0.00 & 0.00 & 0.00 \\
    \bottomrule
  \end{tabularx}
\end{table}

We report the model-graded rubrics and their typed failure rates in Table~\ref{tab:app-quality}. GPT-OSS and Qwen3-30B have much higher attribute-hallucination rates than the other agents while their numeric-hallucination rates remain within the observed range. The aggregate hallucination score alone does not localize this grounding failure.

\begin{table}[h]
  \caption{Model-graded rubrics and failure rates. Rubrics are normalized to $[0,1]$, higher better. Failure rates are per-trial macro-means over the trials to which each applies, lower better.}
  \label{tab:app-quality}
  \centering
  \small
  \setlength{\tabcolsep}{3pt}
  \begin{tabularx}{\textwidth}{@{}Ycccccccc@{}}
    \toprule
    Metric & GPT-OSS & Gem.\,26B & Gem.\,31B & Q3\,30B & Q3.5\,35B & Q3.5\,27B & Q3.6\,35B & Q3.6\,27B \\
    \midrule
    Hallucination           & 0.89 & 0.95 & 0.97 & 0.86 & 0.97 & 0.97 & 0.96 & 0.95 \\
    Response helpfulness    & 0.97 & 0.97 & 0.98 & 0.93 & 0.99 & 0.99 & 0.99 & 0.98 \\
    Coherence               & 0.87 & 0.89 & 0.91 & 0.83 & 0.92 & 0.92 & 0.94 & 0.93 \\
    Tool-call correctness   & 0.70 & 0.76 & 0.76 & 0.72 & 0.74 & 0.76 & 0.73 & 0.76 \\
    Intent understanding    & 0.89 & 0.89 & 0.92 & 0.82 & 0.95 & 0.93 & 0.92 & 0.93 \\
    Clarification           & 0.89 & 0.90 & 0.96 & 0.84 & 0.90 & 0.93 & 0.89 & 0.90 \\
    Recommendation useful  & 0.77 & 0.83 & 0.80 & 0.64 & 0.87 & 0.82 & 0.84 & 0.81 \\
    Add-to-cart behavior    & 0.83 & 0.91 & 0.98 & 0.90 & 0.84 & 0.97 & 0.93 & 0.94 \\
    \midrule
    User frustration        & 0.264 & 0.166 & 0.160 & 0.328 & 0.148 & 0.114 & 0.118 & 0.148 \\
    Phantom action          & 0.060 & 0.081 & 0.076 & 0.158 & 0.055 & 0.084 & 0.060 & 0.050 \\
    Numeric hallucination   & 0.033 & 0.045 & 0.011 & 0.029 & 0.018 & 0.019 & 0.027 & 0.042 \\
    Attribute hallucination & 0.236 & 0.063 & 0.056 & 0.252 & 0.058 & 0.060 & 0.055 & 0.050 \\
    Premature add-to-cart   & 0.153 & 0.063 & 0.018 & 0.050 & 0.146 & 0.026 & 0.066 & 0.051 \\
    Unnecessary clarification     & 0.144 & 0.109 & 0.081 & 0.222 & 0.148 & 0.073 & 0.127 & 0.111 \\
    Failure to clarify      & 0.219 & 0.169 & 0.029 & 0.336 & 0.177 & 0.131 & 0.127 & 0.226 \\
    \bottomrule
  \end{tabularx}
\end{table}

\end{document}